\pdfoutput=1

\documentclass[11pt]{article}

\usepackage[preprint]{acl}

\usepackage{times}
\usepackage{latexsym}

\usepackage[T1]{fontenc}

\usepackage[utf8]{inputenc}

\usepackage{microtype}

\usepackage{inconsolata}

\usepackage{graphicx}

\usepackage{multirow}
\usepackage{rotating}
\usepackage{tabularx}
\usepackage{xurl}
\usepackage{hyperref}
\usepackage{array} 
\usepackage{subcaption}
\usepackage[ruled,vlined,linesnumbered,noend]{algorithm2e}
\SetKwComment{Comment}{$\triangleright$\ }{}
\SetAlgoLined
\usepackage{amsmath}
\newcolumntype{P}[1]{>{\raggedright\arraybackslash}p{#1}} 
\newcolumntype{Y}{>{\centering\arraybackslash}X} 
\usepackage{listings}
\usepackage{listingsutf8} 
\usepackage[table]{xcolor}
\usepackage{booktabs}
\DeclareUnicodeCharacter{2212}{-}

\title{Benchmarking the Benchmarks: Testing the Predictive Validity of Commonsense Benchmarks}
\author{
  \textbf{Ine Gevers} \and \textbf{Walter Daelemans}
\\
  CLiPS, University of Antwerp
\\
  \small{
    \href{mailto:ine.gevers@uantwerpen.be}{ine.gevers@uantwerpen.be},\;
    \href{mailto:walter.daelemans@uantwerpen.be}{walter.daelemans@uantwerpen.be}
  }
}

\begin{document}
\maketitle
\begin{abstract}
Predicting LLM's capabilities on real-world tasks is essential, yet the extent to which performance on commonsense benchmarks predicts downstream performance remains underspecified.
To establish the practical usability of widely adopted commonsense benchmarks, we evaluate 23 models from six families on four established commonsense benchmarks, four reworked variants, three non-commonsense controls, and eight downstream tasks requiring implicit social, pragmatic, temporal, or physical reasoning. We compare model rankings, compute controlled correlations, and use leave-one-family-out cross-validation to assess the criterion validity of commonsense benchmarks.
Our results show that revised benchmarks largely preserve original model rankings and do not improve downstream predictive power. Commonsense benchmarks show consistent cross-family predictive validity for only a narrow subset of downstream tasks, with smaller or metric-specific gains elsewhere.
Overall, standardized commonsense benchmarks provide task-dependent rather than broad evidence of downstream commonsense competence.
\end{abstract}

\section{Introduction}
Standardized evaluation setups allow frequent and relatively cheap model assessments, and such benchmark scores are often interpreted as evidence that a model has acquired a certain skill or knowledge. However, the validity of such claims is questioned (e.g., see \citet{sajadieh2026aiindex,gevers2025benchmarks,sheikhi2026beyond,mitchell2026six}). Many widely used (commonsense) benchmarks are static, multiple-choice datasets, and research has repeatedly shown that such benchmark scores can easily be affected by artifacts, contamination, unstable scoring choices, or unreliable annotations. The central question is no longer whether commonsense benchmarks contain flaws, but whether their scores remain useful indicators of downstream model behavior.\\
We situate this question between two common responses to the current benchmarking crisis. One response repairs and standardizes existing evaluation structures, the other replaces static benchmarks with interactive, task-based, or open-ended evaluation paradigms \citep{truong2026fantastic, kapooropen, ying2026ai, momente2025triangulating}. 
Reworked benchmarks may improve internal dataset quality, but whether these repairs produce more useful indicators of model capability remains unclear. We therefore test both whether revisions preserve the measurements induced by their originals and whether they improve their \textit{criterion validity}: does a model that performs well on a commonsense benchmark also perform well on a downstream task requiring the same underlying skill?
To answer this question, we evaluate 23 LLMs from six model families on four widely used commonsense benchmarks: WinoGrande, HellaSwag, Social IQA, and Physical IQA, and their reworked variants, which modify the originals to address known limitations: WinoWhat, GoldenSwag, a filtered Social IQA variant, and Physical IQA-RUP. The same models are evaluated on eight downstream tasks covering three conceptual axes (social and emotional reasoning, pragmatic inference, and event/physical plausibility): CEI, SARC7, False Beliefs, Implicature, Presupposition, Indirect Requests, TimeDial, and TRIP. 
Because correlations may reflect broad model quality rather than commonsense-specific ability, we include BLiMP, GPQA-Diamond, and MMLU-Redux as controls and test whether commonsense scores improve prediction of models from unseen families beyond model size and control performance.\\ 
The downstream tasks provide a criterion-validity stress test for commonsense benchmarks. 
Solving these tasks requires more than explicitly stated information: models must draw on broadly shared, often unstated (background) knowledge about agents’ beliefs, plausible emotions, speakers’ intentions, and the temporal or physical consequences of events. We do not expect every benchmark to predict every downstream task. However, if a benchmark captures a reusable form of the commonsense reasoning it is intended to measure rather than primarily dataset-specific cues, models that perform well on it should also perform well on contextualized tasks requiring related reasoning. In other words, commonsense benchmarks targeting specific commonsense subskills (e.g., social or physical reasoning) should intuitively transfer better to downstream tasks that are within the same conceptual domain.
Accordingly, we address three research questions: (1) Do commonsense benchmark revisions alter model rankings or improve criterion validity relative to their originals? (2) Do commonsense benchmark scores predict downstream performance beyond general model capability? (3) Is this validity domain-specific, such that social, pragmatic, or event/physical benchmarks are especially predictive of downstream tasks from the same conceptual axis?\\
All tasks are evaluated using a log-likelihood-based multiple-choice setup. We then compute pairwise Spearman correlations, partial Spearman correlations controlling for model size, model family, and non-commonsense benchmark performance, and fit a leave-one-family-out cross-validation predictive model.\\
Our results suggest that first, reworked benchmarks largely preserve the model rankings of their original versions, and do not substantially improve downstream predictive validity. 
Second, False Beliefs and TRIP show the strongest and most consistent associations and cross-family predictive gains. Presupposition also shows positive prediction across metrics, while gains for Implicature, Indirect Requests, and TimeDial are metric-specific; CEI and Sarcasm remain poorly predicted.
Third, we find little evidence that commonsense benchmarks capture distinct domain-specific abilities: transfer rarely follows the fine-grained constructs implied by benchmarks' metadata. 
Together, these findings indicate that commonsense benchmark performance is an uneven predictor of performance on downstream pragmatic tasks. They are most informative for False Beliefs and TRIP and selectively informative elsewhere, but they do not provide broad evidence of pragmatic commonsense ability or reliably support fine-grained claims about distinct commonsense subskills.\footnote{All code and datasets are available at \url{https://github.com/clips/predictive_validity_benchmarks}}

\section{Related Work}
Benchmark scores support capability claims only insofar as the evaluation operationalizes the intended construct. Recent work documents weak discriminability and ranking consistency, sensitivity to prompting and scoring, contamination, and opportunities for metric gaming \citep{qian2026benchmark, molfese2025right, balepur2026benchmarker, chung2026benchmarks, wang2026trustworthybenchmarks}. Relying on benchmarking scores to guide model selections (e.g., as taught in industry tutorials and blogposts: \citet{ibm_llm_benchmarks,arize_llm_leaderboards}) risks misinforming and misdirecting scientific research \citep{bean2026measuring, mitchell2026six}. 
These threats are especially relevant in testing commonsense knowledge, an inherently broad task which suffers from a lack of clear definition, resulting in diverging and broad benchmarks. The observation that commonsense benchmarks are flawed is not new (e.g., see \citet{sajadieh2026aiindex,gevers2025benchmarks,sheikhi2026beyond}). Their often constrained, multiple-choice question answer format may not reflect their practical model behaviour \citep{balepur2025these,wang2024my}; answer labels can clash with human preferences or vary across populations \citep{palta2024plausibly,nguyen2026large}; and inherent subjective tasks can lead to unstable annotations \citep{kejriwal2024noise}.\\
Dataset cleaning can address some internal weaknesses, but improved annotation quality does not by itself establish criterion validity. The distinction between internal and criterion validity is particularly relevant for revised benchmarks. Revisions are frequently justified by showing that examples are less ambiguous, less artifact-driven, or more robust to a particular shortcut. These are important improvements, but they are therefore not automatically better indicators of downstream performance. A reworked benchmark can remain highly correlated with its original because filtering preserves the same relative advantages across models; alternatively, extensive rewriting can change rankings while measuring a different mixture of abilities. Comparing original and revised benchmarks against external criteria makes these possibilities empirically distinguishable, a research gap that this study addresses.\\
Our study aims to empirically establish the predictive validity of standardized commonsense benchmarks: instead of focusing on the ideal response to current benchmarking flaws, we explore the actual relationship between academic benchmark scores and downstream utility, which remains relatively unmapped \citep{sheikhi2026beyond}. This moves evaluation from internal dataset quality to \textit{criterion validity}: whether benchmark-induced differences between models relate to relevant external outcomes, and incremental validity: whether they add information beyond general-capability indicators.

Commonsense reasoning is generally understood not as a single, formally delimited skill, but as a broad body of everyday background knowledge concerning physical objects, events, social relations, beliefs, goals, emotions, norms, and likely consequences \citep{mccarthy1959programs,davisMarcus2015commonsense,levesque2012winograd}. 
Pragmatic interpretation relies on much of the same implicit information: implicatures, presuppositions, indirect speech, sarcasm, and socially situated interpretations require reasoning about speaker goals, common ground, context, and plausible events \citep{grice1975logic,stalnaker2002commonGround,geurts2024commonGround,sravanthi2024pub,ma2025pragmaticsSurvey}.

Nevertheless, prior work indicates that performance does not necessarily transfer even across benchmarks intended to measure related capabilities \citep{shen2023experimental}. Recent studies have consequently examined capability relationships through cross-model correlations and predictive comparisons \citep{fan2026magic,tsvilodub2026emergent}. However, the extent to which standardized commonsense benchmarks predict performance on more contextualized pragmatic and commonsense-dependent tasks remains insufficiently studied. We address this gap by testing whether the model rankings induced by widely used commonsense benchmarks transfer to downstream tasks requiring related forms of implicit reasoning.

\section{Methodology}
\subsection{Commonsense benchmarks}
We select four widely used commonsense benchmarks and four revisions designed to filter problematic items or paraphrase the original data (see Table~\ref{tab:original_benchmarks}). All are evaluated on public validation splits using length-normalized answer-option log likelihood.

The selected benchmarks operationalize complementary aspects of the broad commonsense construct. WinoGrande requires resolving ambiguous references through knowledge about events, goals, and causal relations; HellaSwag asks models to distinguish plausible from implausible continuations of everyday situations. Social IQA focuses on intentions, motivations, reactions, and likely social consequences, while Physical IQA targets object use, affordances, and intuitive physical constraints. Despite these different emphases, all assume that successful answering requires recovering information not explicitly stated in the prompt.

\begin{table*}[t]
\centering
\footnotesize
\setlength{\tabcolsep}{3pt}
\renewcommand{\arraystretch}{1.02}

\caption{Commonsense benchmark pairs \citep{sakaguchi2021winogrande,zellers2019hellaswag,sap2019socialiqa,bisk2020piqa,gevers2025winowhat,chizhov2025hellaswag,mousavi2026garbage,wang2024rupbench}.}
\label{tab:original_benchmarks}

\begin{tabular}{
    @{}
    >{\raggedright\arraybackslash}p{3.1cm}
    >{\raggedright\arraybackslash}p{4.2cm}
    >{\raggedright\arraybackslash}p{4.2cm}
    >{\raggedright\arraybackslash}p{1.7cm}
    >{\centering\arraybackslash}p{1.65cm}
    @{}
}
\toprule
\textbf{Original} &
\textbf{Target} &
\textbf{Reworked} &
\textbf{Revision} &
\textbf{n options} \\
\midrule

\mbox{WinoGrande (2019)} \newline
$n \approx 1{,}000$
&
Coreference resolution
&
\mbox{WinoWhat (2025)} \newline
$n \approx 1{,}000$
&
Paraphrased
&
2 options
\\

\mbox{HellaSwag (2019)} \newline
$n \approx 10{,}000$
&
Event and narrative plausibility
&
\mbox{GoldenSwag (2025)} \newline
$n \approx 1{,}500$
&
Filtered
&
4 options
\\

\mbox{Social IQA (2019)} \newline
$n \approx 2{,}000$
&
Social interactions, intentions, and reactions
&
\mbox{Social IQA filtered (2026)} \newline
$n \approx 1{,}300$
&
Filtered
&
3 options
\\

\mbox{Physical IQA (2020)} \newline
$n \approx 2{,}000$
&
Physical commonsense and affordances
&
\mbox{Physical IQA RUP (2024)} \newline
$n \approx 1{,}800$
&
Paraphrased
&
2 options
\\

\bottomrule
\end{tabular}
\end{table*}

The availability of both the original and revised benchmarks allows us to measure whether benchmark revisions (1) affect model rankings and (2) are more indicative of downstream performance compared to their original counterpart.

\subsection{Downstream tasks}
We construct a downstream evaluation suite spanning social inference, pragmatic interpretation, discourse understanding, and event-level reasoning. These tasks were selected because they instantiate the implicit reasoning targeted by commonsense benchmarks in richer linguistic and discourse contexts. 
We organize the tasks along three conceptual axes (see Table \ref{tab:downstream}). The \textbf{social and emotional reasoning} axis comprises CEI \citep{chun2026cei}, SARC7 \citep{xiong2025sarc7}, and False Beliefs \citep{jones2024comparing}. These tasks concern reasoning about agents’ internal states: their emotions, attitudes, intentions, and representations of the world. The \textbf{pragmatic-inference} axis instead concerns meaning conveyed through language but not fully determined by its literal content. It comprises Implicature and Presupposition \citep{jeretic2020natural}, and Indirect Requests \citep{jones2024comparing}, which require models to reason about communicative intentions, conversational alternatives, and common ground. The \textbf{event and physical plausibility} axis comprises TimeDial \citep{qin2021timedial} and TRIP \citep{storks2021tiered}, which require temporal event reasoning, physical state tracking, and inference about the preconditions and consequences of actions.\\ 
These axes are analytical groupings rather than a definitive taxonomy of commonsense. They capture broad correspondences between the intended skills of the standardized benchmarks and the forms of reasoning required downstream: Social IQA relates to the social and emotional axis; HellaSwag and Physical IQA relate most directly to event and physical plausibility; and all four commonsense benchmarks may contribute to pragmatic inference because pragmatic interpretation can draw jointly on social, intentional, causal, and physical knowledge.

We manually removed ambiguous, near-duplicate, unnatural, or contextually insufficient examples where required (see Appendix \ref{app:analysis_details}). This yields small evaluation sets for some tasks, but prioritizes interpretable labels over scale.  We summarize the main statistics of the downstream tasks in Table \ref{tab:downstream}. 

\begin{table}[t]
\centering
\small
\setlength{\tabcolsep}{4pt}
\renewcommand{\arraystretch}{1.02}
\begin{tabularx}{\columnwidth}{@{}p{2.25cm}Y@{}}
\toprule
\textbf{Axis} & \textbf{Downstream tasks} \\
\midrule
Social/emotional & CEI ($n=31$); SARC7 ($n\!\approx\!100$); False Beliefs ($n\!\approx\!192$) \\
Pragmatic & Indirect Requests ($n\!\approx\!64$); Presupposition ($n\!\approx\!201$); Implicature ($n\!\approx\!121$) \\
Event/physical & TimeDial ($n=60$); TRIP ($n\!\approx\!100$) \\
\bottomrule
\end{tabularx}
\caption{Downstream tasks grouped by conceptual axis.}
\label{tab:downstream}
\end{table}

Similarly to the standardized benchmarks, we evaluate the downstream tasks using averaged log-likelihood. While it has been established that such metric has flaws (e.g., see \citet{boseak2025evaluating}), it is the most consistent way to keep the evaluation format in our study uniform across heterogeneous tasks. It is important to point out that the goal of our study is not to improve model performance, but we are interested in the correlation between model scores on different benchmarks, for which using the same evaluation setup is indispensable.
For CEI, Implicature, Presupposition, and SARC7, whose labels repeat across instances, we append a short answer statement (e.g., `the speaker feels [label]') and calibrate label priors following \citet{zhao2021calibrate}:
\begin{equation}
\mathrm{Score}(y)=\log p(y\mid x)-\log p(y\mid \text{N/A}).
\end{equation}

\subsection{Control benchmarks}
To distinguish commonsense-specific transfer from broad model capability, we include three multiple-choice QA benchmarks that rely on factual knowledge/ general language knowledge. Specifically, we include subsets from BLiMP (adjunct island, anaphor number agreement, superlative quantifiers) \citep{warstadt2020blimp}; MMLU-Redux (abstract algebra, formal logic, and college math) \citep{gema2025we}; and GPQA Diamond \citep{rein2023gpqa}.

\subsection{Model selection}
We evaluate 23 open-weight models from Llama 3, Gemma 2, Pythia, OPT, Qwen 2.5, and Mistral, spanning 1B–72B parameters. Model specifics are listed in Appendix \ref{appendix-models}.

\subsection{Comparing original and reworked commonsense benchmarks} \label{methodology:validity}
We first examine whether benchmark revisions change how models are ranked, independently of their relationship with downstream tasks. For each matched original--reworked pair, we compute the Spearman correlation across models.

We then test whether revisions systematically alter the broader correlation structure among commonsense benchmarks. We classify pairwise correlations as original--original (OO), reworked--reworked (RR), or original--reworked (OR). Correlations are Fisher-$z$ transformed before being averaged within each category. We compare the mean OO correlation with the RR and OR means to assess whether reworked benchmarks remain as mutually consistent with the broader benchmark set as the originals.

Uncertainty is estimated using a nonparametric bootstrap over models, analogous to the Boot-Systems procedure of \citet{deutsch2021statistical}. In each iteration, models are sampled with replacement, all pairwise Spearman correlations are recomputed and Fisher-$z$ transformed, and the group-level differences are calculated. The resulting distributions provide 95\% confidence intervals while preserving the dependence among benchmark scores for each model. This analysis therefore establishes whether revisions preserve or disrupt the model-ranking structure of the original benchmarks. Section~\ref{sec:criterion_validity} subsequently examines the separate question of whether either benchmark version transfers more strongly to downstream tasks.

\subsection{Criterion validity tests} \label{sec:criterion_validity}
We treat downstream performance as a criterion-validity test of standardized commonsense benchmarks. Rather than claiming that any individual downstream task defines commonsense reasoning, we ask whether the model rankings induced by commonsense benchmarks transfer to tasks requiring related forms of implicit inference.

This cross-model approach follows prior work showing that strong in-domain performance does not necessarily transfer across commonsense reasoning benchmarks \citep{shen2023experimental}. Related studies use correlations over model accuracy and confidence to test whether benchmark rankings remain stable across training or evaluation settings \citep{fan2026magic}. Closest to our study, \citet{tsvilodub2026emergent} use behavioral correlations and predictive comparisons to examine relationships between theory-of-mind and pragmatic-reasoning tasks. We adopt the same general logic, but apply it to the relationship between standardized commonsense benchmarks and downstream pragmatic tasks.

If commonsense benchmarks measure reusable reasoning abilities, models that perform well on them should also perform well on downstream tasks requiring inference about events, beliefs, intentions, emotions, affordances, and temporal relations. We additionally test whether transfer is domain-specific: Social IQA should preferentially predict social and emotional tasks, while HellaSwag- and Physical IQA-style benchmarks should preferentially predict event and physical plausibility tasks. The pragmatic axis is treated as broader and exploratory because pragmatic interpretation may draw on several forms of commonsense. Weak or non-specific transfer would instead suggest that benchmark scores primarily reflect dataset-specific regularities or broad model capability.

As a baseline, we compute pairwise Spearman correlations between each commonsense benchmark and each downstream task. This rank-based measure aligns well with our use case: if benchmark scores are used to compare or select models, the central question is whether high-scoring models on a commonsense benchmark are also ranked highly on downstream tasks.

To test whether reworked benchmarks provide stronger criterion validity than their originals, we compare their respective Spearman correlations with each downstream task \(Y\):
\begin{equation}
\Delta\rho
=
\rho(\mathrm{Original},Y)
-
\rho(\mathrm{Reworked},Y).
\end{equation}
Positive values indicate stronger criterion validity for the original benchmark, whereas negative values favor its reworked counterpart. We estimate 95\% confidence intervals for each difference using paired bootstrap resampling over models, following the procedure introduced in Section~\ref{methodology:validity}.

Because raw correlations may reflect general model capability rather than commonsense-specific skill, we also compute partial Spearman correlations. We rank-transform the benchmark and downstream scores, residualize both with respect to log model size, model family, and performance on non-commonsense control benchmarks, and correlate the resulting residuals.

We assess domain specificity using the three conceptual axes introduced above. For each axis, we compare Fisher-\(z\)-transformed correlations for same-axis and cross-axis benchmark--task pairs. We also test whether in-axis commonsense benchmarks predict a downstream axis better than out-of-axis benchmarks and conduct a global comparison of all axis-matched and mismatched pairs. Uncertainty is estimated through paired bootstrap resampling over models. 

Finally, we evaluate out-of-sample validity using leave-one-family-out cross-validation (LOFOCV). Ridge regression is trained on five families and predicts each model in the unseen sixth family. We compare log model size plus the non-commonsense composite against models adding the original commonsense composite, the reworked composite, or both, using pooled cross-validated \(R^2\) and Spearman correlation. With six family folds, results are descriptive. Tests use two-sided Benjamini--Hochberg FDR correction \citep{benjamini1995controlling}. Analyses use accuracy and macro-F1, except TimeDial, whose macro-F1 is uninformative because the correct answer always occupies the same position. Appendix~\ref{app:analysis_details} provides details.

\section{Results and Discussion}

\subsection{Revisions preserve model rankings and do not improve criterion validity}
\begin{figure}[t]
    \centering
    \includegraphics[width=0.85\linewidth]{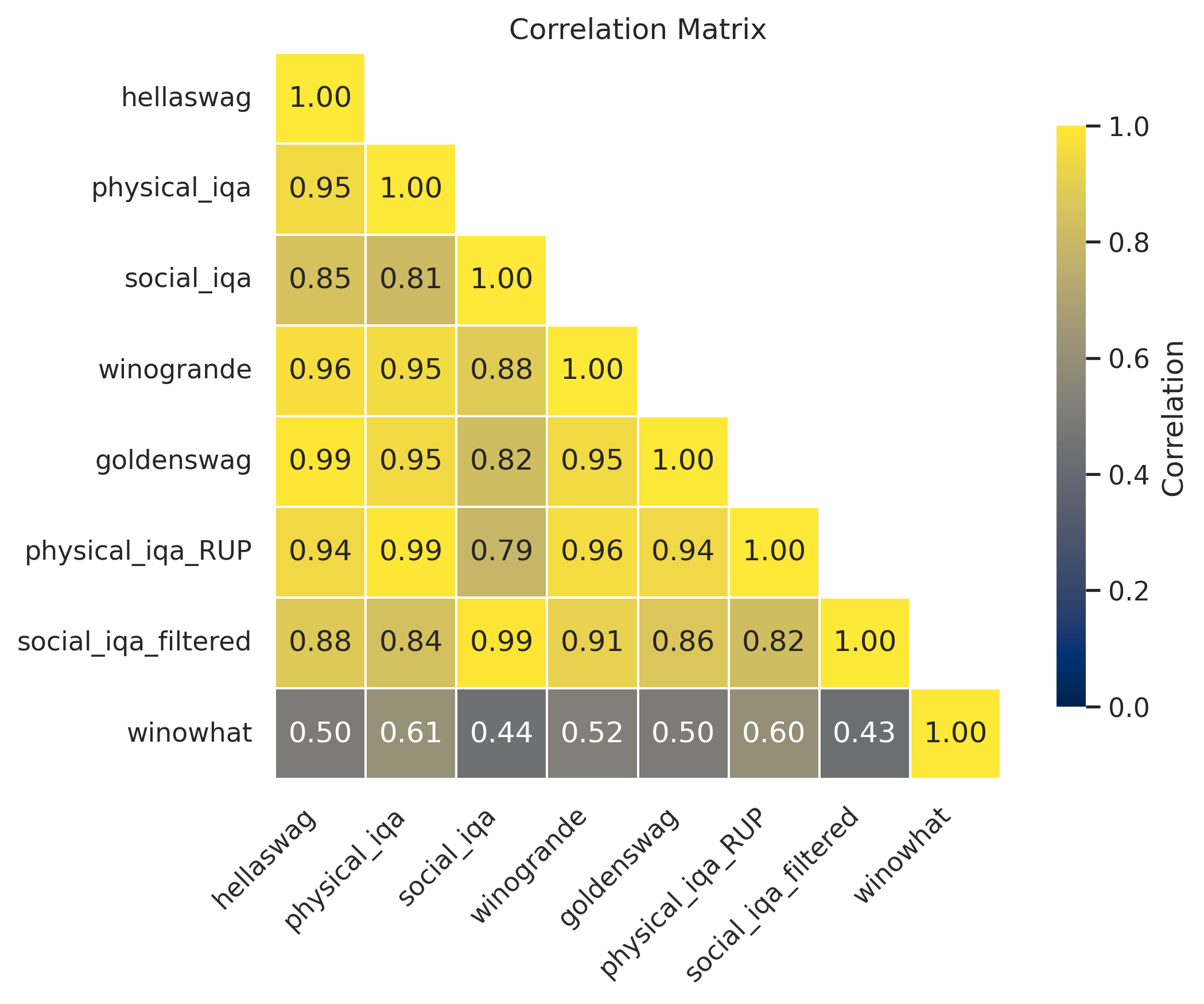}
    \caption{Correlation matrix heatmap for original and reworked common sense benchmarks.}
    \label{fig:correlation_heatmap}
\end{figure}

Figure~\ref{fig:correlation_heatmap} shows the Spearman correlation matrix across the original and reworked commonsense benchmarks. Generally, reworked benchmarks preserve model rankings, except Winogrande - WinoWhat: WinoWhat correlates less strongly with WinoGrande and the other commonsense benchmarks. In practical terms, most reworked benchmarks do not strongly alter which models appear better or worse; WinoWhat is the only case where the revision appears to capture meaningfully different model behavior.

The group-level correlation analysis confirms this interpretation. Here, we test whether revisions systematically change the model-ranking structure by comparing mean Fisher-$z$ correlations within the original and reworked benchmark groups. Originals are more strongly intercorrelated than revisions ($\bar{z}=1.565$ vs.\ $.982$; $\Delta=.584$, 95\% CI $[.326,.853]$), and this difference remains positive when each model family is excluded. However, the weaker correlations between original and reworked benchmarks appear to be largely attributable to WinoWhat, rather than to benchmark revisions generally.

Further, we test whether revisions predict downstream performance better compared to their original counterparts. Across downstream tasks, most bootstrap intervals comparing original and reworked correlations include zero, indicating no reliable downstream-validity advantage for reworked benchmarks.
The clearest difference is in the opposite direction: WinoGrande is substantially more predictive of False Beliefs than WinoWhat ($\rho_{\mathrm{OG}}=.883$, $\rho_{\mathrm{RW}}=.351$, $\Delta\rho=.532$). For HellaSwag--GoldenSwag, Physical IQA--Physical IQA-RUP, and Social IQA--Filtered, we find no reliable evidence that reworked benchmarks improve downstream prediction.

This result is important because benchmark revisions are often motivated by the assumption that cleaner or less artifact-driven datasets should provide more valid measurements. Our findings do not show such an improvement at the level of downstream predictive validity. In most cases, reworked benchmarks preserve the same model-ranking information as the original benchmark; in the WinoGrande--WinoWhat case, the revision changes the ranking structure but does not improve downstream prediction. Thus, reworking benchmarks may address dataset-level flaws, but this does not automatically translate into stronger criterion validity.

\subsection{Downstream transfer is narrow}
\begin{figure}[t]
    \centering
    \includegraphics[width=0.85\linewidth]{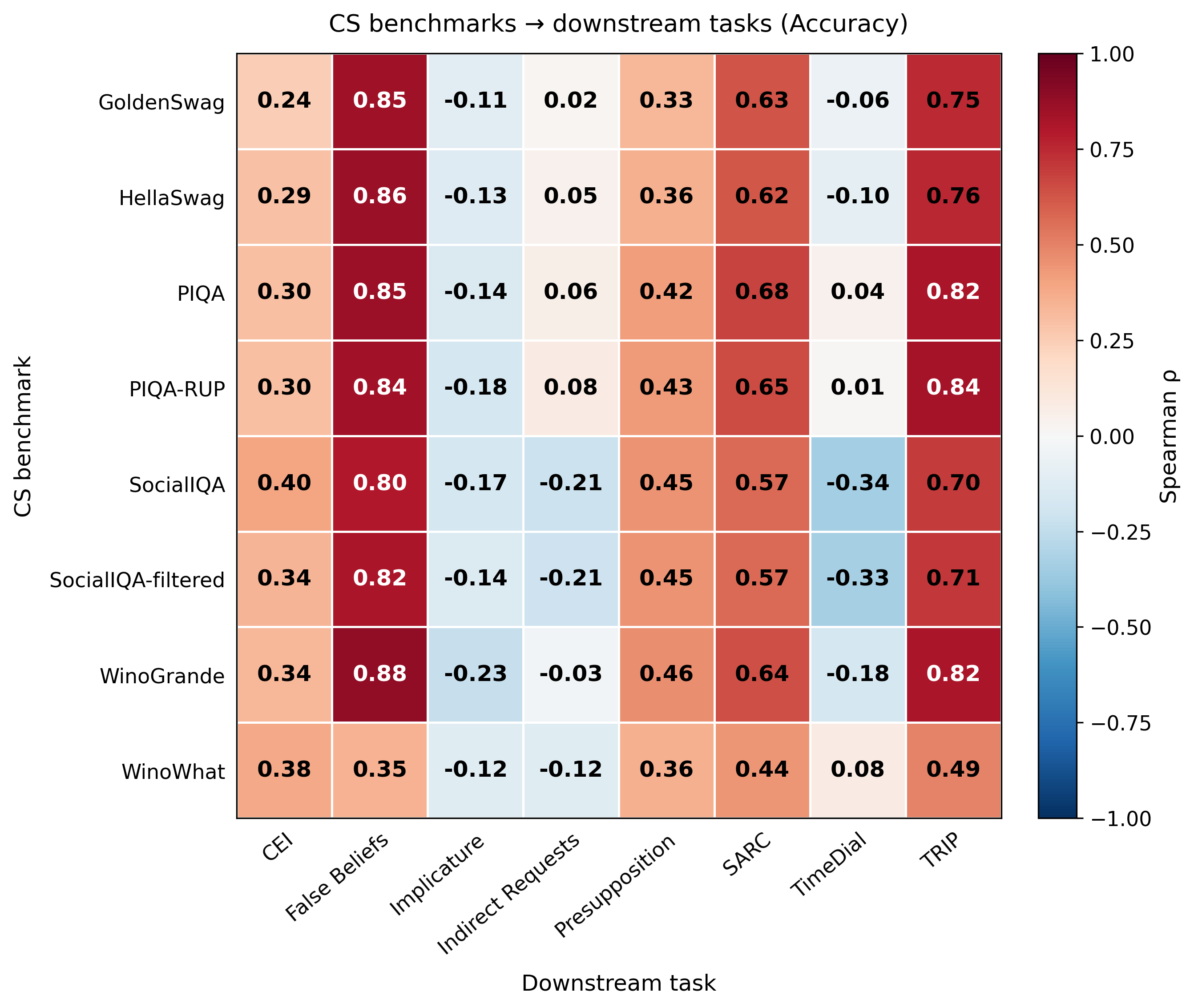}
    \caption{Correlation matrix heatmap for commonsense benchmarks and downstream tasks.}
    \label{fig:correlation_heatmap_cs-downstream}
\end{figure}

Figure \ref{fig:correlation_heatmap_cs-downstream} shows the Spearman correlations between each commonsense benchmark and each downstream task. We observe that overall, commonsense benchmarks transfer to a narrow subset of downstream tasks, rather than to pragmatics broadly.

Significant positive correlations are concentrated on False Beliefs and TRIP. SARC7 also shows broad significant positive transfer for accuracy, although this pattern is less consistent across metrics. By contrast, no positive correlations with CEI, Presupposition, TimeDial, or Indirect Requests survive FDR correction. Implicature shows the clearest inverse pattern: several negative macro-F1 correlations remain significant after FDR correction, with negative associations also emerging for Indirect Requests. Overall, significant transfer is therefore highly task-dependent and largely restricted to False Beliefs, TRIP, and, to a lesser extent, SARC7. 

This narrow transfer has direct implications for model selection: models performing best on commonsense benchmarks are not necessarily the best models for downstream tasks. Across commonsense--downstream comparisons, the top-three models overlap never exceeds an average Jaccard score of .25, while models shift by six rank positions on average. Thus, performance on commonsense benchmarks does not reliably identify the best-performing models across the downstream tasks evaluated here (see Appendix~\ref{app:model_rank_displacement}).

\subsection{Predictive validity generalizes across families for selected tasks}
The narrow transfer pattern raises a further question: are commonsense benchmarks predictive because they measure commonsense-specific abilities, or because they track broader model capability? To address this, we  compute partial Spearman correlations controlling for model size, model family, and control non-commonsense benchmark performance.

The control benchmarks show that downstream prediction is not unique to commonsense benchmarks. Non-commonsense controls also exhibit significant associations with individual downstream tasks, with MMLU-Redux notably predicting False Beliefs and BLiMP predicting TimeDial accuracy. This suggests that some of the observed commonsense--downstream correlations may reflect broader model capabilities rather than commonsense-specific criterion validity.

After controlling for model size, model family, and non-commonsense benchmark scores, no extended partial Spearman correlation survives FDR correction. Some estimates remain large, especially for HellaSwag--TRIP, but the adjusted $p$-values are non-significant. The controlled analysis therefore provides insufficient evidence that commonsense benchmarks explain downstream variation beyond broader model capability, although its statistical power is constrained by the number of models and controls.

\begin{figure}[t]
    \centering
    \includegraphics[width=0.85\linewidth]{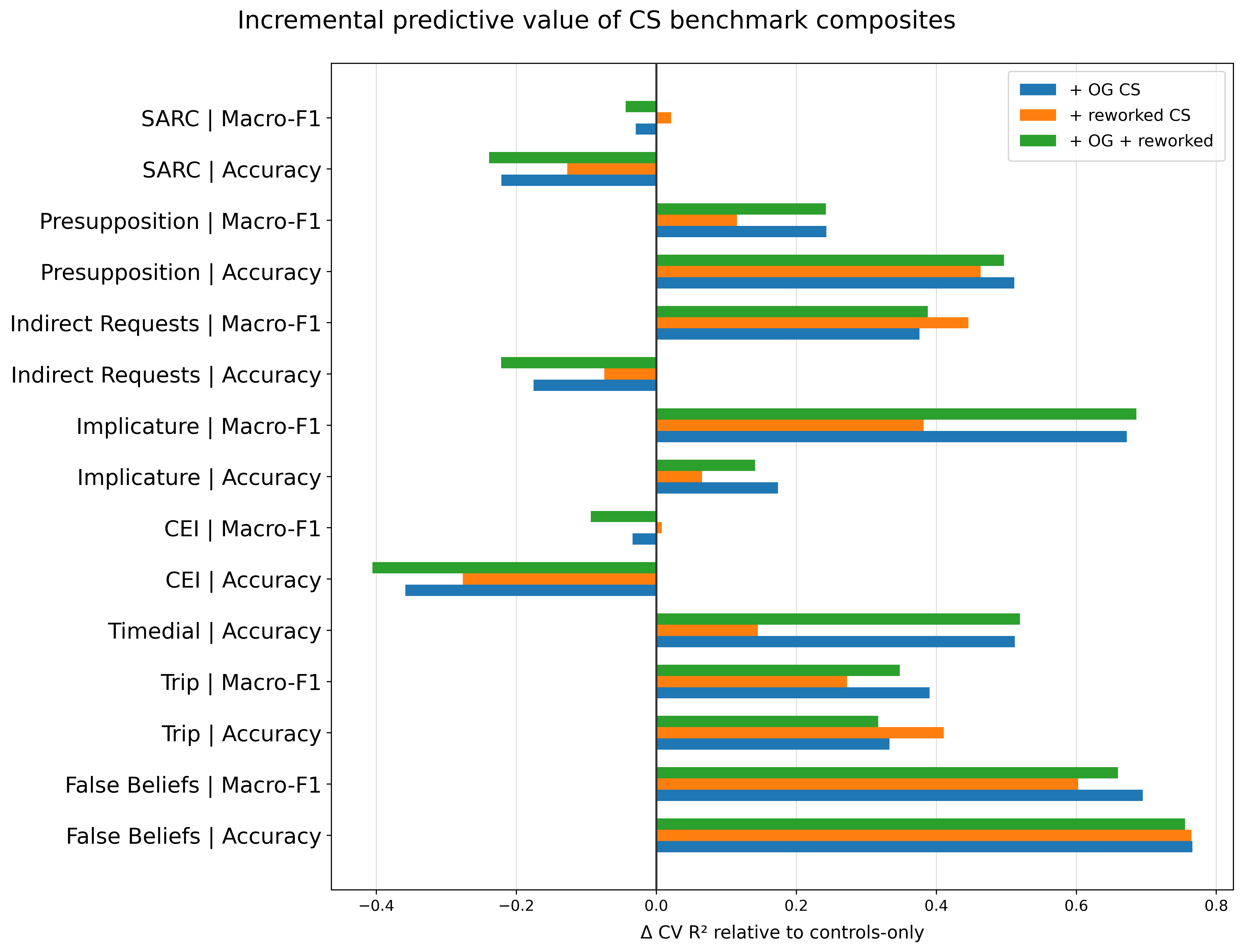}
    \caption{Change in LOFOCV $R^2$ when CS composites are added to controls. Positive values indicate improved prediction for models from unseen families.}
    \label{fig:lofocv_delta_vs_controls}
\end{figure}

LOFOCV tests practical prediction for unseen families. Figure \ref{fig:lofocv_delta_vs_controls} shows the strongest, most consistent gains for False Beliefs and TRIP. Their best commonsense-augmented models reach \(R^2=.483/.271\) and \(.451/.397\) for accuracy/macro-F1, improving over controls by \(.766/.695\) and \(.411/.390\), respectively. Presupposition is also positive across metrics (\(R^2=.242/.077\)). Implicature and Indirect Requests are positive only for macro-F1, and TimeDial only for accuracy, while all CEI and Sarcasm models have negative \(R^2\). Commonsense scores therefore contain cross-family information, but it remains task- and metric-dependent.

Further, original composites are not systematically inferior to revisions, and combining both rarely improves on the better single composite. This mirrors the pairwise analysis: the revisions mostly preserve the predictive signal already available in the originals. The lack of consistent gains across the suite limits the scope of the corresponding criterion validity claim.

We hypothesize that False Beliefs and TRIP may transfer better because they are conceptually and operationally closer to standardized commonsense evaluation: both require selecting the alternative consistent with an agent's belief state or a sequence of physical state changes. TRIP especially resembles HellaSwag and Physical IQA. In contrast, Implicature, Presupposition, and Indirect Requests depend more heavily on subtle discourse context and speaker intentions. This interpretation remains tentative: attenuation under controls suggests broad capability contributes, while the LOFOCV gains show that commonsense scores retain criterion-relevant information that generalizes across families.
 
\begin{figure*}[t]
    \centering

    \begin{subfigure}[t]{0.48\textwidth}
        \centering
        \includegraphics[width=\linewidth]{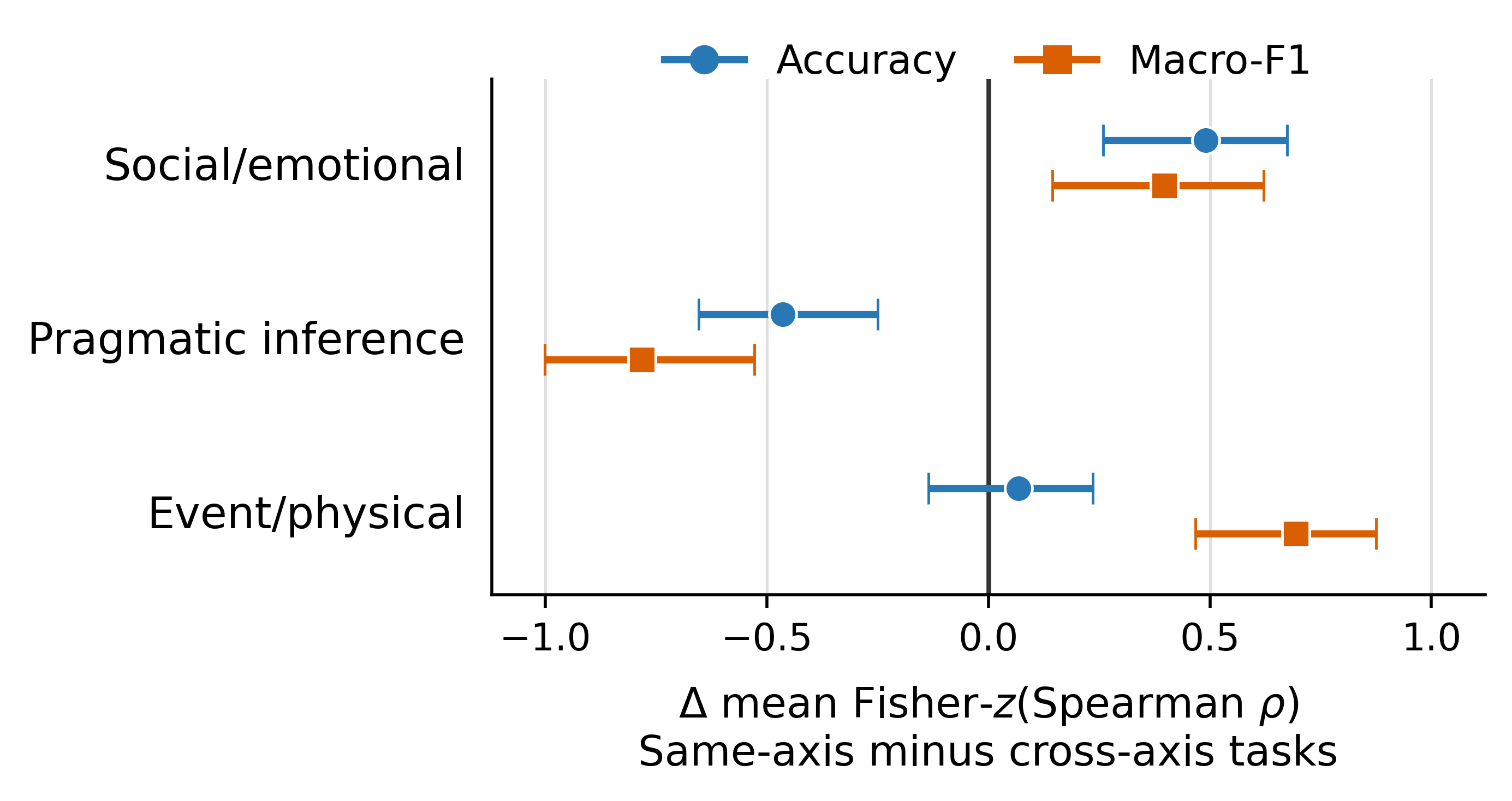}
        \caption{Same-axis versus other downstream tasks.}
        \label{fig:axis_specificity}
    \end{subfigure}
    \hfill
    \begin{subfigure}[t]{0.48\textwidth}
        \centering
        \includegraphics[width=\linewidth]{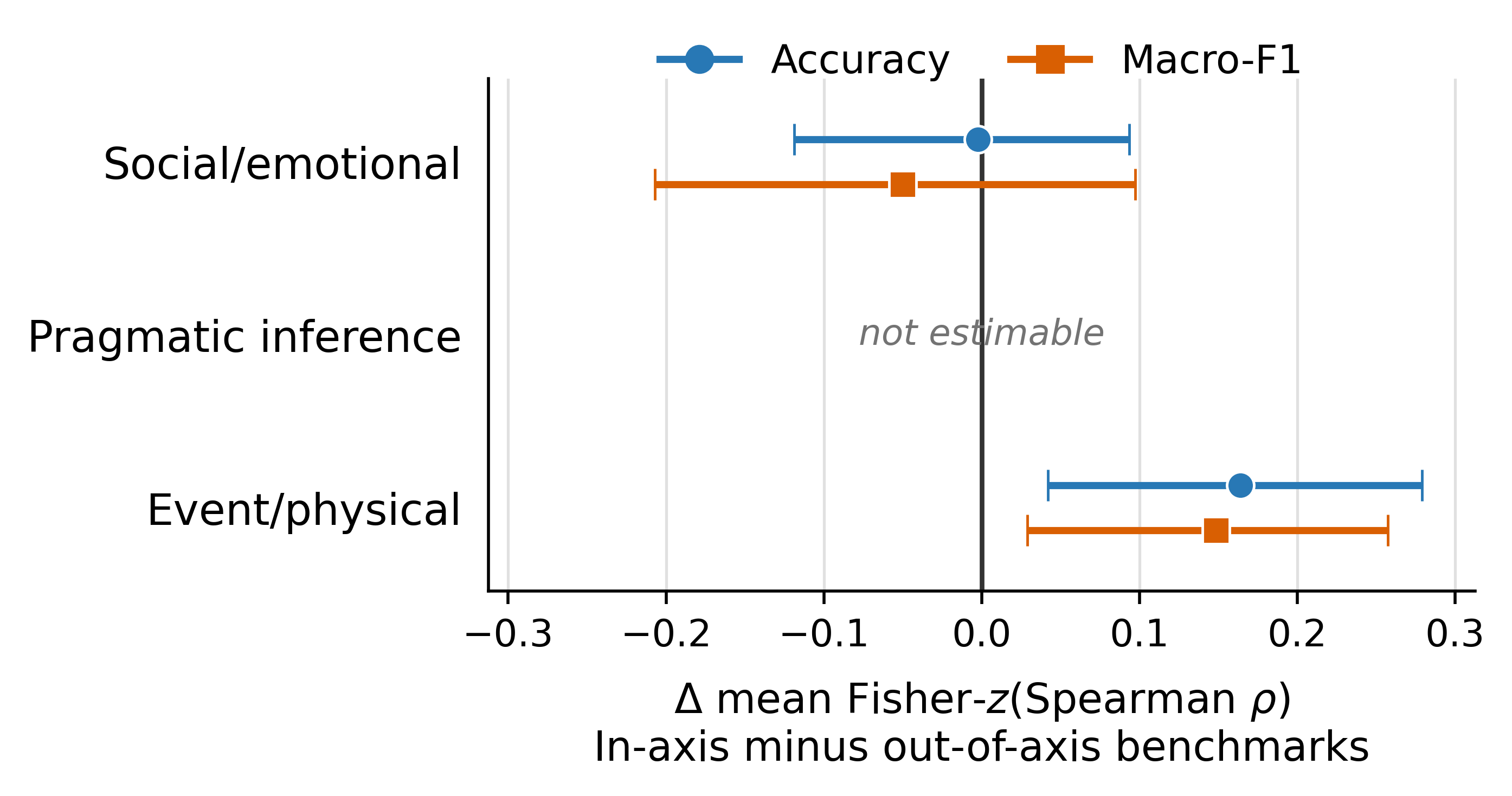}
        \caption{Same-axis versus other CS benchmarks.}
        \label{fig:axis_downstream}
    \end{subfigure}

    \caption{
    Axis-specific criterion-validity tests. Panel (a) compares same- and cross-axis downstream prediction; panel (b) compares same- and cross-axis CS predictors. Points show differences in mean Fisher-$z$-transformed Spearman correlations, with 95\% bootstrap CIs; positive values indicate axis alignment. Pragmatic inference is omitted from panel (b) because no out-of-axis CS comparison is available.
    }
    \label{fig:axis_validity}
\end{figure*}

\subsection{Downstream validity is not cleanly axis-specific}
If benchmark labels reflect distinct abilities, axis-matched benchmarks and downstream tasks should correlate more strongly than mismatched pairs. Figure~\ref{fig:axis_validity} tests this in two directions: panel (a) compares prediction of same- versus other-axis downstream tasks, while panel (b) compares same- versus other-axis commonsense benchmarks.

Evidence for this pattern is limited. The pragmatic axis shows the opposite pattern for both metrics. The social axis shows an advantage only in panel (a), while event/physical plausibility provides the clearest, though not fully consistent, alignment. Globally, matched pairs are weaker than mismatched pairs for both metrics. Thus, transfer to particular tasks does not reliably follow the domains implied by benchmark labels, limiting fine-grained capability claims.

\section{Conclusion}
This study shifts the benchmark-validity debate from whether commonsense benchmarks contain artifacts to whether their scores are useful predictors of downstream tasks. Across 23 models from six families, we examined whether benchmark revisions change model rankings, whether commonsense scores transfer to downstream tasks, whether this transfer is specific to the intended commonsense domain, and whether it remains informative beyond broader model capability.

Our central finding is that standardized commonsense benchmarks have task-dependent and uneven downstream usefulness. Benchmark rankings generalize most consistently to False Beliefs and TRIP and improve Presupposition prediction more modestly across both metrics. Gains for Implicature, Indirect Requests, and TimeDial are metric-specific, while CEI and SARC7 remain poorly predicted. 
Commonsense benchmark performance therefore should not be interpreted as broad evidence of pragmatic or contextual reasoning ability, and its relationships rarely align with the subdomains implied by benchmark labels.

Our results also challenge the assumption that repairing a benchmark necessarily improves its criterion validity. Most reworked benchmarks preserve model rankings compared to their original counterparts, and none of them systematically improve downstream prediction. Dataset cleaning and artifact reduction may therefore improve benchmark quality without making the resulting scores more useful for model selection or capability inference.

These findings question the capability-level interpretation that underlies much commonsense benchmarking. Performance on HellaSwag, Social IQA, Physical IQA, or WinoGrande provides evidence about performance on that benchmark, but does not by itself measure broad commonsense competence. Extending a leaderboard gain into claims about pragmatic understanding, social reasoning, or downstream robustness requires evidence of criterion validity: scores must generalize to external tasks that instantiate the intended capability in different contexts and formats, and contribute information beyond broader model ability. The same standard should apply to benchmark revisions, which should demonstrate not only improved annotation quality or reduced artifacts, but stronger external and incremental criterion validity. The priority for commonsense benchmarking should therefore shift from repeatedly refining isolated static datasets toward building evaluation suites whose scores are demonstrably informative about the behaviors and applications they are intended to represent.

\section{Limitations}
Our 23 models span six families and 1B–72B parameters, providing broad coverage across model scales and lineages. Nevertheless, the number of models and families limits the precision of analyses that simultaneously account for size, family, and non-CS performance. We therefore interpret the controlled correlations and six-fold LOFOCV results jointly, focusing on consistent patterns across analyses rather than individual estimates.

Our downstream suite samples social, pragmatic, temporal, and physical reasoning but does not exhaust the possible manifestations of commonsense. The proposed axes serve as theory-informed analytical groupings, and some evaluation sets are relatively small after qualitative filtering. The conclusions should therefore be understood as applying to the tasks and capability dimensions examined here.

Finally, we restricted evaluation to option log-likelihoods because it provides a uniform, prompt-invariant, and deterministic metric across heterogeneous tasks. Unlike free-form generation, which introduces severe evaluator bias, decoding hyperparameter noise, and prompt sensitivity, calibrated log-likelihoods (Eq. 1) isolate model representation quality under identical decision constraints. Future work can test whether the observed validity patterns extend across evaluation formats and to application-level outcomes.

\section*{Acknowledgments}
This research was made possible with a grant from the Fonds Wetenschappelijk Onderzoek (FWO) project 11P3824N.

\bibliography{custom}

\appendix
\section{Model specifications} \label{appendix-models}
In Table \ref{tab:model_specifications}, we detail model name, version, and release date of the models included in our experiments. Because of hardware limitations, we ran the largest models (>12B) on rented GPU instances through Vast.ai\footnote{\url{https://vast.ai/}}; models up to ~32B parameters were evaluated on a single NVIDIA RTX PRO 6000 Blackwell GPU (96GB VRAM), while larger models (OPT-66B, Llama-3.3-70B-Instruct, Qwen2.5-72B-Instruct) were evaluated on a dual-GPU configuration (2× NVIDIA RTX PRO 6000, 96GB VRAM each, 192GB total). The total cost for prompting the models was 65\$.

\begin{table*}[t]
\centering
\scriptsize
\setlength{\tabcolsep}{3pt}
\renewcommand{\arraystretch}{0.95}

\caption{Models included in the evaluation. Parameter counts are nominal
sizes reported by the model developers. Pythia and OPT are base models;
all other checkpoints are instruction-tuned.}
\label{tab:model_specifications}

\begin{tabularx}{\textwidth}{
    @{}
    >{\raggedright\arraybackslash}p{2.2cm}
    >{\raggedright\arraybackslash}p{3.0cm}
    >{\raggedright\arraybackslash}X
    @{}
}
\toprule
\textbf{Family} &
\textbf{Parameter sizes} &
\textbf{Hugging Face checkpoints} \\
\midrule

Pythia (2023)
&
1B, 2.8B, 6.9B, 12B
&
\path{EleutherAI/pythia-\{1b,2.8b,6.9b,12b\}}
\\

OPT (2022)
&
1.3B, 2.7B, 6.7B, 13B, 66B
&
\path{facebook/opt-\{1.3b,2.7b,6.7b,13b,66b\}}
\\

Qwen 2.5 (2024)
&
1.5B, 3B, 7B, 14B, 72B
&
\path{Qwen/Qwen2.5-\{1.5B,3B,7B,14B,72B\}-Instruct}
\\

Llama 3 (2024)
&
1B, 3B, 8B, 70B
&
\path{meta-llama/Llama-3.2-\{1B,3B\}-Instruct};
\path{Llama-3.1-8B-Instruct};
\path{Llama-3.3-70B-Instruct}
\\

Gemma 2 (2024)
&
2B, 9B, 27B
&
\path{google/gemma-2-\{2b,9b,27b\}-it}
\\

Mistral (2024)
&
7B, 12B
&
\path{mistralai/Mistral-7B-Instruct-v0.3};
\path{mistralai/Mistral-Nemo-Instruct-2407}
\\

\bottomrule
\end{tabularx}
\end{table*}

\section{Additional analysis details} \label{app:analysis_details}
\paragraph{Filtering and prompts.}
CEI examples with ambiguous labels and SARC7 examples lacking sufficient context were removed. Near-duplicates and unnatural or unclear instances were removed from IMPPRES and TRIP. TimeDial contexts were shortened to place MASK sentence-finally where possible. EPITOME False Beliefs and Indirect Requests were not filtered.

\paragraph{Axis-based analysis.}
For each conceptual axis, we compare the Fisher-$z$ mean Spearman correlation for axis-matched benchmark--task pairs with the corresponding mean for axis-mismatched pairs. Differences are evaluated using paired bootstrap resampling over models with 5,000 iterations. We additionally perform a global comparison between all benchmark--task pairs that share at least one axis and those that do not. Positive differences indicate stronger prediction for conceptually aligned pairs. The pragmatic axis is treated as exploratory because pragmatic inference may draw jointly on social, event-level, and physical commonsense; consequently, all commonsense benchmarks are considered potentially relevant to this axis.

\paragraph{Out-of-sample prediction.}
We construct original, reworked, and non-CS composites by averaging their constituent benchmark scores after standardization. We compare four ridge-regression predictor sets: log model size plus the non-CS composite; controls plus the original CS composite; controls plus the reworked CS composite; and controls plus both CS composites. The $L_2$ penalty is fixed at $\alpha=1$. In each LOFOCV fold, all models from one family are withheld, while composite standardization, imputation, feature scaling, and model fitting use only the other five families. Training-fold means and standard deviations are then applied to the held-out family, and every held-out model is predicted individually. Performance is calculated over the 23 pooled out-of-family predictions using cross-validated $R^2$ and Spearman correlation; we additionally inspect family-balanced MAE and RMSE. Because there are only six independent family folds, results are interpreted descriptively.

\paragraph{Statistical testing.}
All significance tests are two-sided. We apply the Benjamini--Hochberg false discovery rate correction within each predefined family of comparisons \citep{benjamini1995controlling}. Unless otherwise stated, analyses are conducted for both accuracy and macro-F1. TimeDial is evaluated using accuracy only because all correct responses occupy the same output class, making macro-F1 uninformative.

\section{Model rank displacements} \label{app:model_rank_displacement}
To complement the pairwise correlation analysis, we examine the practical stability of model rankings when moving from commonsense benchmarks to downstream tasks. For each benchmark–task pair, we compute the Spearman rank correlation, the mean absolute displacement of models in the ranking, and the Jaccard overlap between the top-three models. Table~\ref{tab:rank_instability_downstream} reports these measures averaged across downstream tasks. The results show limited ranking preservation: models typically move by approximately six to seven positions, and the overlap among the three highest-ranked models remains low. This analysis is descriptive, but it illustrates that benchmark rankings may offer limited guidance for selecting models for downstream use.

\begin{table}[t]
\centering
\scriptsize
\setlength{\tabcolsep}{3.5pt}
\renewcommand{\arraystretch}{1.05}

\caption{Correspondence between model rankings on each commonsense
benchmark and the downstream tasks. Accuracy averages over
eight downstream tasks; macro-F1 averages over seven, excluding TimeDial.}
\label{tab:rank_instability_downstream}

\begin{tabular}{@{}lccc@{}}
\toprule
\textbf{CS benchmark} &
\(\boldsymbol{\overline{\rho}}\) &
\textbf{Mean absolute} &
\textbf{Mean top-3} \\
&
&
\textbf{rank displacement} &
\textbf{Jaccard} \\
\midrule

\multicolumn{4}{@{}l}{\textit{Accuracy}} \\
Physical IQA          & .379 & 5.65 & .213 \\
Physical IQA-RUP      & .373 & 5.67 & .213 \\
WinoGrande            & .343 & 5.73 & .250 \\
HellaSwag             & .340 & 5.90 & .100 \\
GoldenSwag            & .335 & 5.93 & .163 \\
Filtered Social IQA   & .282 & 6.27 & .200 \\
Social IQA            & .279 & 6.28 & .200 \\
WinoWhat              & .250 & 6.25 & .138 \\

\addlinespace[2pt]
\multicolumn{4}{@{}l}{\textit{Macro-F1}} \\
Physical IQA          & .206 & 6.35 & .200 \\
Physical IQA-RUP      & .202 & 6.35 & .200 \\
WinoGrande            & .181 & 6.51 & .243 \\
HellaSwag             & .141 & 6.72 & .143 \\
Social IQA            & .133 & 6.98 & .200 \\
GoldenSwag            & .131 & 6.79 & .157 \\
Filtered Social IQA   & .125 & 6.99 & .200 \\
WinoWhat              & .078 & 7.26 & .029 \\

\bottomrule
\end{tabular}
\end{table}

\end{document}